\documentclass[lettersize,journal]{IEEEtran}
\usepackage{amsmath,amsfonts}
\usepackage{algorithmic}
\usepackage{algorithm}
\usepackage{array}
\usepackage[caption=false,font=footnotesize,labelfont=rm,textfont=rm]{subfig}
\usepackage{textcomp}
\usepackage{stfloats}
\usepackage{url}
\usepackage{verbatim}
\usepackage{graphicx}
\usepackage{cite}
\usepackage{amssymb}
\usepackage{multirow}
\usepackage{makecell}
\begin{document}
	
	\title{Offline and Online Optical Flow Enhancement for \\ Deep Video Compression}
	
	\author{Chuanbo Tang, Xihua Sheng, Zhuoyuan Li, Haotian Zhang, Li Li,~\IEEEmembership{Member,~IEEE}, Dong Liu,~\IEEEmembership{Senior Member,~IEEE}
		\thanks{C. Tang, X. Sheng, Z. Li, H. Zhang, L. Li and D. Liu are with the CAS Key Laboratory of Technology in Geo-spatial Information Processing and Application System, Department of Electronic Engineering and Information Science, University of Science and Technology of China, Hefei, 230027, China (e-mail: cbtang@mail.ustc.edu.cn; xhsheng@mail.ustc.edu.cn; zhuoyuanli@mail.ustc.edu.cn; zhanghaotian@mail.ustc.edu.cn; lil1@ustc.edu.cn; dongeliu@ustc.edu.cn).}
		\thanks{Corresponding author: Dong Liu.}
	}
	
	
	
	\maketitle
	
	\begin{abstract}
		Video compression relies heavily on exploiting the temporal redundancy between video frames, which is usually achieved by estimating and using the motion information. The motion information is represented as optical flows in most of the existing deep video compression networks. Indeed, these networks often adopt pre-trained optical flow estimation networks for motion estimation. The optical flows, however, may be less suitable for video compression due to the following two factors. First, the optical flow estimation networks were trained to perform inter-frame prediction as accurately as possible, but the optical flows themselves may cost too many bits to encode. Second, the optical flow estimation networks were trained on synthetic data, and may not generalize well enough to real-world videos. We address the twofold limitations by enhancing the optical flows in two stages: offline and online. In the offline stage, we fine-tune a trained optical flow estimation network with the motion information provided by a traditional (non-deep) video compression scheme, e.g. H.266/VVC, as we believe the motion information of H.266/VVC achieves a better rate-distortion trade-off. In the online stage, we further optimize the latent features of the optical flows with a gradient descent-based algorithm for the video to be compressed, so as to enhance the adaptivity of the optical flows. We conduct experiments on a state-of-the-art deep video compression scheme, DCVC. Experimental results demonstrate that the proposed offline and online enhancement together achieves on average 12.8\% bitrate saving on the tested videos, without increasing the model or computational complexity of the decoder side.
	\end{abstract}
	
	
	\section{Introduction}
	To meet the demand for video transmission and storage, video compression has been developed for several decades. Video compression relies heavily on exploiting the temporal redundancy between video frames, which is usually achieved by estimating and using the motion information. The motion information is represented as optical flows in most of the existing deep video compression networks~\cite{lu2019dvc,lu2020end,li2021deep,sheng2022temporal,lin2020m,shi2022alphavc,li2022hybrid, li2023neural}. Indeed, these networks often adopt pre-trained optical flow estimation networks~\cite{ranjan2017optical,ilg2017flownet,sun2018pwc} to estimate the motions between video frames. Taking a widely acknowledged and highly flexible scheme, DCVC~\cite{li2021deep}, as an example, the pre-trained Spynet~\cite{ranjan2017optical} is used for estimating the optical flows. The optical flows can be considered as pixel-wise motion vectors (MV) and are compressed by an autoencoder-based MV encoder~\cite{minnen2018joint}. In the training stage, the pre-trained Spynet is first loaded, and then the whole deep video compression network is optimized in an end-to-end manner. In the inference stage, the motion information of different video contents is obtained through the fixed networks.
	
	However, regarding the optical flows estimated by the commonly-used pre-trained optical flow estimation networks~\cite{ranjan2017optical,ilg2017flownet,sun2018pwc} as motion information in deep video compression schemes may be less suitable due to the following two factors. First, the pre-trained optical flow estimation networks are trained to perform inter-frame prediction as accurately as possible, but the optical flows themselves may cost too many bits to encode. Although they can be further optimized with the whole video compression networks in an end-to-end manner, the inappropriate initial point may affect the final optimization result. Second, the optical flow estimation networks are trained on synthetic data~\cite{dosovitskiy2015flownet,butler2012naturalistic,baker2011database}, and may not generalize well enough to real-world videos. The end-to-end optimization in video compression networks can alleviate the domain gap between the synthetic data and the real-world videos to some degree. However, once the end-to-end optimization is finished, the optical flow estimation network is "optimal" in the sense that the average performance over the entire training set is optimal, but not "optimal" in the sense that the network produced optical flows may not be the optimal for any given video sequence.
	
	
	To address the twofold limitations, we consider learning the good traditions from the inter-frame prediction techniques in traditional (non-deep) video compression schemes. The latest traditional video compression standard H.266/VVC~\cite{bross2021developments} has achieved great success in effectively estimating and using the motion information, which is represented by MV. Specifically, in the offline stage, various hand-crafted inter-frame prediction modes are first designed for different types of motions without optimization. Then, the optimal mode is searched online to achieve the best rate-distortion (RD) performance for each coding sequence. Such offline and online optimization is believed a promising direction for learning-based video compression as well in the reference~\cite{huo2022towards}.
	
	Similar to the two-stage strategy in the traditional video compression scheme, we address the twofold limitations of the optical flows by enhancing them in two stages: offline and online. In this paper, we propose an offline and online enhancement on the optical flows to better estimate and utilize motion information under the RD constraint. Specifically, in the offline stage, the trained optical flow estimation network Spynet is fine-tuned by the MV provided by VTM (reference software of H.266/VVC), as we believe the MV of VTM achieves a better RD trade-off. With the guidance of the MV of VTM, the optical flow estimation network can provide a more appropriate initial point for end-to-end optimization in video compression networks.
	In terms of the online stage, we optimize the latent features of the optical flows with a gradient descent-based algorithm for the video to be compressed, so as to enhance the adaptivity of the optical flows. Inspired by the search-based online optimization algorithm in traditional video compression schemes, our scheme enables online updating the latent features of the optical flows by minimizing the RD loss in the inference stage, which has been introduced in deep image compression~\cite{campos2019content}. When online updating the latent features of the optical flows, the parameters of the whole video compression networks are fixed and the decoding time remains unchanged. With the online enhancement, the updated latent features can help the video compression networks achieve a better RD performance than the latent features obtained by a simple forward pass through the MV encoder. 
	
	We conduct experiments on the widely acknowledged baseline DCVC~\cite{li2021deep} to verify the effectiveness of our scheme. Experimental results demonstrate our scheme can outperform DCVC without increasing the model size or computational complexity on the decoder side. It is worth noting that our scheme is a plug-and-play mechanism that can be easily integrated into any deep video compression framework with the same motion estimation and MV encoder. 
	
	Our contributions are summarized as follows:              
	\begin{itemize}
		\item We propose an offline enhancement on the optical flows by fine-tuning the optical flow estimation network with the MV of VTM. With the guidance of the MV of VTM, the optical flow estimation network can provide a more appropriate initial point for end-to-end optimization in deep video compression networks.
		\item We further enhance the adaptivity of the optical flows by online optimizing the latent features of the optical flows according to the contents of different coding sequences in the inference stage without changing the network parameters.
		\item When equipped with our proposed offline and online optical flow enhancement methods, the baseline scheme DCVC achieves a better RD performance without increasing the model size and decoding complexity.
	\end{itemize}
	
	\section{RELATED WORK}
	
	\subsection{Deep Video Compression}
	Recently, deep video compression has explored a new direction for video compression.
	The mainstream of deep video compression frameworks can be divided into two categories: the motion-compensated prediction and residual coding framework and the motion-compensated prediction and conditional coding framework. 
	DVC~\cite{lu2019dvc} is the pioneering work for the motion-compensated prediction and residual coding framework, which
	followed the traditional video compression framework and replaced all the modules with neural networks. Different from the residual coding-based framework, DCVC\cite{li2021deep} introduced the motion-compensated prediction and conditional coding framework, which is able to utilize the learned temporal correlation between the current frame and predicted frame rather than the subtraction-based residual. Many existing works have followed these two frameworks and conducted research on the frameworks and optimization strategy.
	
	{\bfseries Research on the frameworks.} For the motion-compensated prediction and residual coding framework, many existing works~\cite{lu2020end,lin2020m,hu2020improving,hu2021fvc,hu2022coarse,shi2022alphavc} have followed this framework to improve the compression ratio. Lu $et\ al.$~\cite{lu2020end} utilized the auto-regressive model~\cite{minnen2018joint} to compress the motion and residual, and added the motion and residual refine modules to further improve the compression performance. Lin $et\ al.$~\cite{lin2020m} proposed multi-frame-based motion estimation and motion compensation to reduce the temporal redundancy efficiently. The deformable convolution~\cite{dai2017deformable} was applied for motion estimation, compression, and compensation in the feature domain\cite{hu2021fvc}. In \cite{hu2020improving}, the single-resolution motion compression was extended to resolution-adaptive motion compression in both the frame level and block level. Hu $et\ al.$~\cite{hu2022coarse} proposed coarse-to-fine motion compensation to reduce the residual energy and hyperprior-guided adaptive motion and residual compression to realize the block partition and residual skip. Shi $et\ al.$~\cite{shi2022alphavc} has improved the compression performance by introducing the conditional-I frame, pixel-to-feature motion prediction, and probability-base entropy skipping method.
	
	Following the motion-compensated prediction and conditional coding framework, Sheng $et\ al.$~\cite{sheng2022temporal} further proposed the multi-scale temporal context mining to better utilize the temporal correlation. The following work \cite{li2022hybrid} designed a parallel-friendly entropy model which explores both temporal and spatial dependencies. Li $et\ al.$~\cite{li2023neural} further increased the context diversity in both temporal and spatial dimensions by introducing the group-based offset diversity and quadtree-based partition.
	
	Most existing deep video compression schemes often adopt pre-trained optical flow estimation networks~\cite{ranjan2017optical,ilg2017flownet,sun2018pwc} for motion estimation.
	However, the optical flow estimation networks are trained on synthetic data~\cite{dosovitskiy2015flownet,butler2012naturalistic,baker2011database} to only perform inter-frame prediction as accurately as possible without considering the bits cost on the estimated optical flows. In comparison, in our offline enhancement stage, the pre-trained optical flow estimation network is fine-tuned by the MV of VTM, which is extracted from real-world videos and achieves a better RD trade-off than the optical flows.
	
	\begin{figure*}[t]
		\centering
		\includegraphics[width=0.94\textwidth]{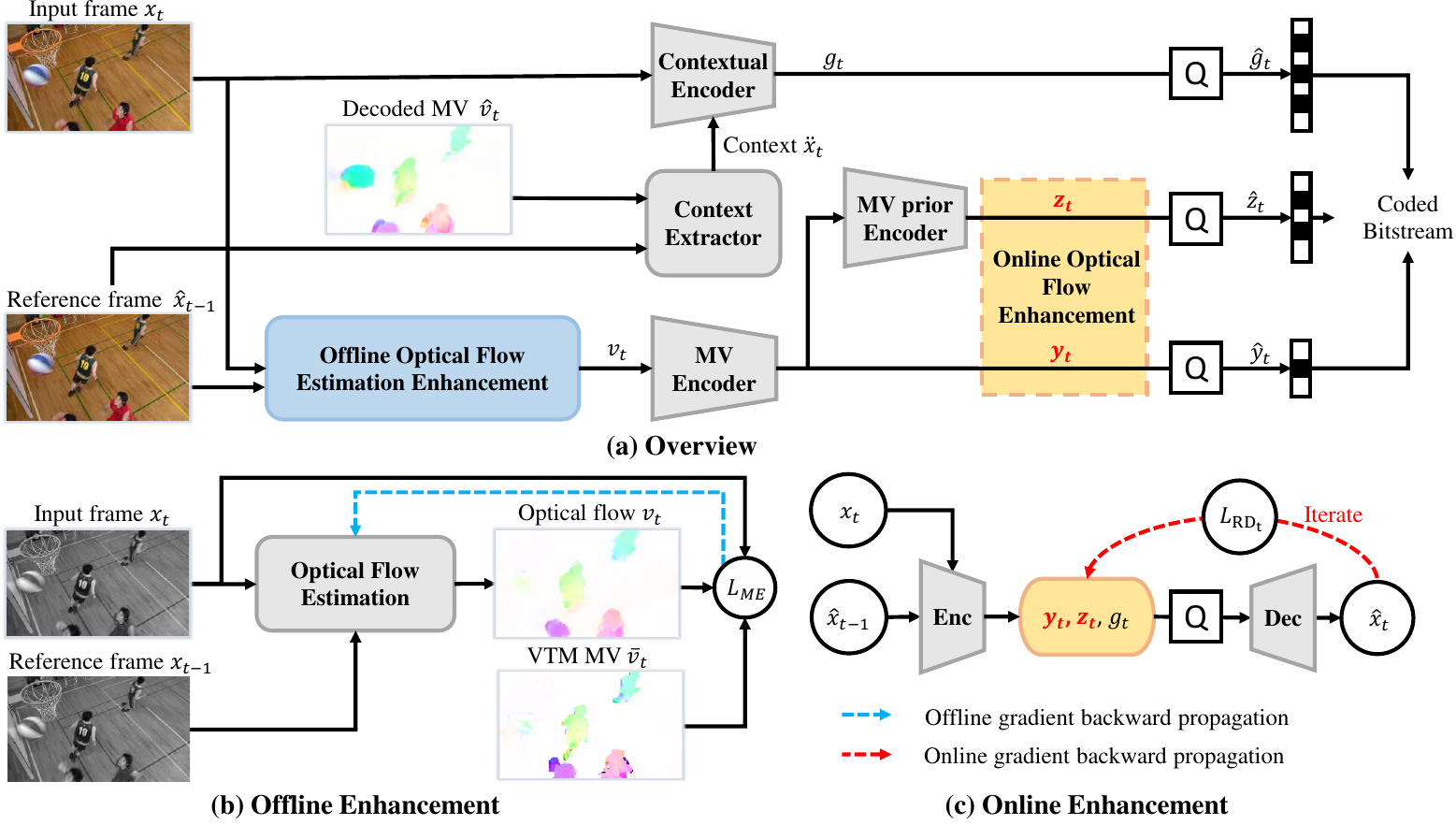}
		\caption{(a) Overview of our framework. (b) Offline enhancement for the optical flow network. The training procedure of the optical flow estimation network is supervised by the MV of VTM. (c) Online enhancement for the optical flow. The latent features of the optical flow $y_t$ and $z_t$ are online updated by minimizing the RD loss $L_{RD_t}$.}
		\label{overview}
	\end{figure*}
	
	{\bfseries Research on the optimization strategy.} Lu $et\ al.$~\cite{lu2020content} applied a new training objective with multiple time steps to prevent the error accumulation and adopted an online encoder updating scheme to realize the content adaptive. The online encoder updating scheme needs to update the parameters of the encoder while keeping the parameters of the decoder fixed in the inference stage. BAO~\cite{xu2022bit} is a pixel-level implicit bit allocation method for deep video compression by using iterative optimization. Xu $et\ al.$~\cite{xu2022correcting} proved the BAO's sub-optimality and recursively applied back-propagating on non-factorized latent through gradient ascent-based on the corrected optimal bit allocation algorithm. 
	
	In contrast, as video compression relies heavily on estimating and using the motion information to exploit the temporal redundancy between video frames, in our online enhancement stage, we only optimize the latent features of the optical flows with a gradient descent-based algorithm.

	\subsection{Inter-frame Prediction in Traditional Video Compression}
	Traditional video compression have been developed for several decades, and many video compression standards have been proposed, such as H.264/AVC~\cite{wiegand2003overview}, H.265 /HEVC~\cite{sullivan2012overview}, and H.266/VVC~\cite{bross2021developments}. These video compression schemes follow a similar hybrid video coding framework, where inter-frame prediction techniques play a crucial role among the techniques.
	
	In the latest coding standard (H.266/VVC~\cite{bross2021developments}), many advanced inter-frame prediction techniques have been proposed to attain high inter-frame coding efficiency. To estimate the accurate motion, various motion situations (translation, rotation motion model, etc.) corresponding to different inter-frame prediction modes (AMVP~\cite{chien2021motion}, Affine~\cite{li2017efficient}, etc.) are executed to search for the optimal MV for each coding region via Rate-Distortion-Optimization (RDO)~\cite{sullivan1998rate}. For each coding sequence, the optimal mode is searched online from multiple inter-frame prediction modes. With the increase in the number of inter-frame prediction modes and the ever-rising searching cost, traditional video compression frameworks have achieved great compression performance.
	
	Considering the latest traditional video codec VTM~\cite{bross2021developments} searches each MV for the best RD performance for each coding sequence, the MV can achieve a better RD trade-off than the optical flows. Thus, in this paper, we enhance the optical flows in two stages according to the inter-frame prediction techniques in the traditional video compression scheme. In the offline stage, we propose an offline enhancement on the optical flows by fine-tuning the trained optical flow estimation network Spynet~\cite{ranjan2017optical} with the MV searched by VTM. In the online stage, inspired by the VTM searching-based strategy in motion estimation, we further optimize the latent features of the optical flows with a gradient descent-based algorithm while the parameters of the whole network are fixed. Our strategy is to search the optimal latent features of the optical flows for the best RD performance by varying only the latent values themselves.
	
	\section{PROPOSED METHOD}
	\subsection{Overview}
	In this paper, our proposed offline and online enhancement is integrated into the baseline scheme DCVC\cite{li2021deep} to demonstrate the effectiveness. The encoding procedure of our scheme, as illustrated in Fig.~\ref{overview}(a), can be divided into three parts: motion estimation, motion compression, and contextual compression.
	
	{\bfseries Motion Estimation.} The input frame $x_t$ and the reference frame $\hat{x}_{t-1}$ are fed into our proposed offline enhanced optical flow estimation network to estimate the optical flows, which are considered as pixel-wise MV $v_t$. Following DCVC, the network is based on Spynet\cite{ranjan2017optical}, but we fine-tune it with the MV of VTM. Details are presented in Section~\ref{off_enh}.
	
	{\bfseries Motion Compression.} The estimated MV $v_t$ is compressed by an autoencoder-based MV encoder~\cite{minnen2018joint}. The latent features of the optical flows, MV features $y_t$ and MV hyperprior $z_t$, are online enhanced by updating with a gradient descent-based algorithm in the inference stage.
	More information is provided in Section~\ref{on_enh}.
	
	{\bfseries Contextual Compression.} Following DCVC, the input frame $x_t$ is compressed conditioned on the context $\ddot{x}_t$, which is extracted by the context extractor using the reference frame $\hat{x}_{t-1}$ and the decoded MV $\hat{v}_{t}$ as input. 
	
	\subsection{Offline Enhancement} \label{off_enh}
	To alleviate the domain gap between the synthetic data and the real-world videos, and provide a more appropriate initial point for the end-to-end optimization in deep video compression networks, we propose the offline enhancement on the optical flows. Different from DCVC, we fine-tune the pre-trained Spynet with the MV searched by VTM for the best RD performance on real-world videos, which has a better RD trade-off than the optical flows. 
	
	{\bfseries Preliminaries.} To provide the optical flow estimation network with accurate and learnable labels, we extract the block-level MV from each frame by VTM under certain configuration. To match the low-delay mode of DCVC, the reference list of VTM is set to only include the previous frame of the current frame. Besides, for acquiring finer MV on the encoder side, we set the quantization parameter (QP) to 22 and turn off the decoder-side MV refine technique (PROF~\cite{luo2019prediction}). As the coding block predicted by intra mode is not appropriated for the training of optical flow estimation network, we turn off the intra-prediction mode and intra-related inter technique (combine inter-intra prediction, CIIP~\cite{chien2021motion}) in VTM to obtain the MV. The extracted block-level MV is at the quarter resolution, so we use the nearest interpolation to obtain the full-resolution block-level MV $\overline{v}_t$ and scale the upsampled MV by a factor of 16.
	
	Specifically, as shown in Fig. \ref{overview}(b), we fine-tune the pre-trained Spynet under the guidance of the extracted MV~$\overline{v}_t$, which is searched by VTM for the best RD performance on real-world videos. 
	To better match the warp operation in the video compression, our training objective for Spynet is to minimize both the End Point Error (EPE) loss and the Mean Squared Error (MSE) loss between the input frame and the corresponding warp frame. Let $\check{x_t}$ denote the warp frame, 
	\begin{equation} 
		\label{warp}
		\setlength{\abovedisplayskip}{3pt}
		\setlength{\belowdisplayskip}{3pt}
		\check{x}_t = w(x_{t-1}, v_t), 
	\end{equation}
	where $w(\cdot)$ denotes the warp operation. Therefore, our training objective is to minimize a weighted sum of EPE and MSE loss,
	\begin{equation} 
		\label{epe_mse}
		\setlength{\abovedisplayskip}{3pt}
		\setlength{\belowdisplayskip}{3pt}
		{L_{ME} = \frac{1}{mn}\sum_{i,j}\sqrt{(v_i - \overline{v}_i)^2 + (v_j - \overline{v}_j)^2} + \lambda_{ME} \cdot d(x_t, \check{x}_t)}.
	\end{equation}
	The $m \times n$ in Eq. (\ref{epe_mse}) is the image dimension and the $i$ and $j$ subscript indicate the horizontal and vertical components of the flow vector and motion vector. $d(x_t, \check{x_t})$ represents the MSE metric for measuring the difference between the input frame $x_t$ and the warp frame $\check{x_t}$. $\lambda_{ME}$ is the Lagrange multiplier that controls the trade-off between the EPE and MSE loss.
	
\begin{figure}
		\centering
		\includegraphics[width=\linewidth]{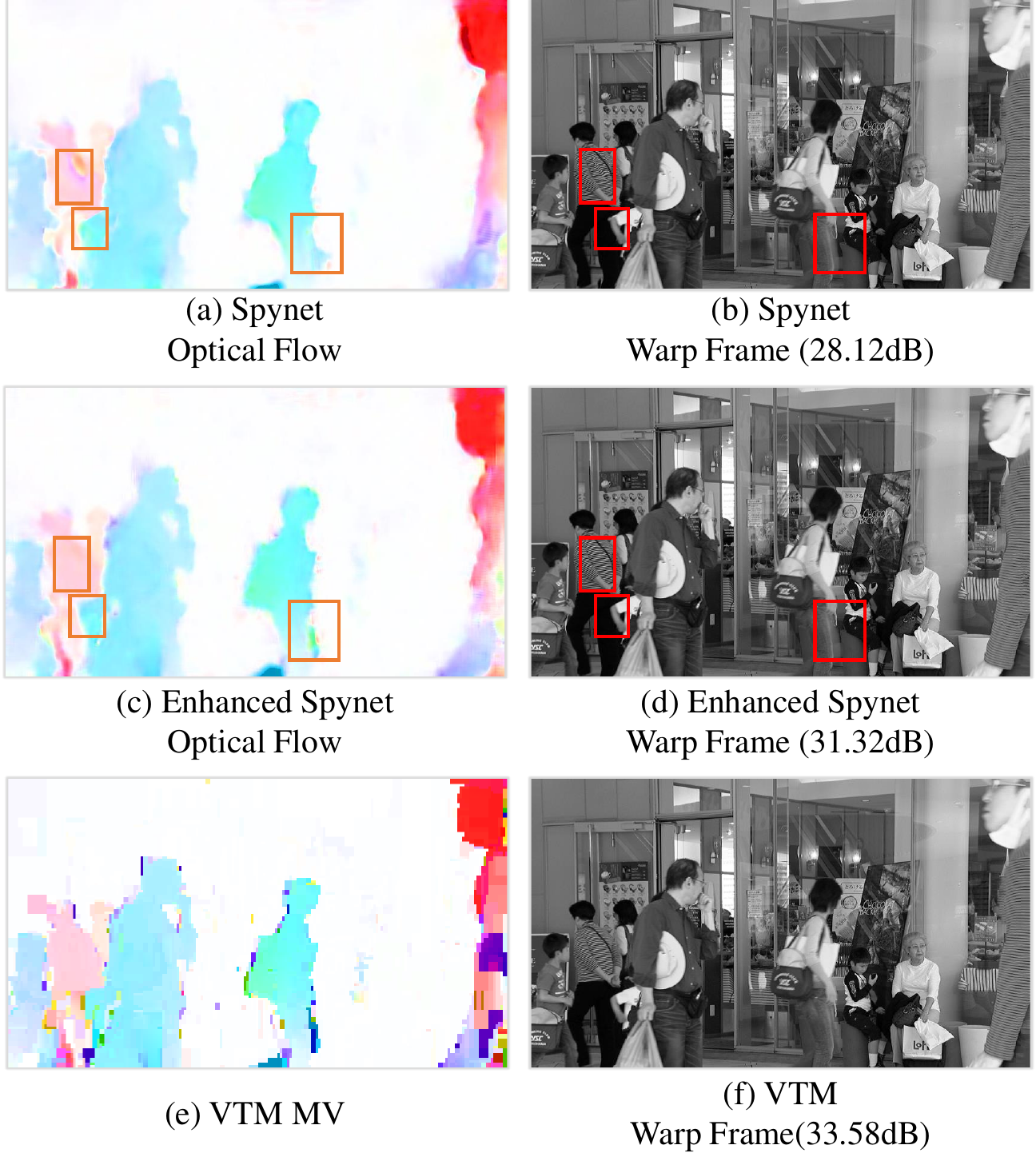}
		\caption{ Visualization of the optical flow estimated by Spynet (a) and enhanced Spynet (c), VTM MV (e), and the corresponding warp frame (b, d, f) for the 29th frame in the BQMall\_832$\times$480\ sequence from JVET CTC Test Class C dataset. }
		\label{visual_offline}
		\vspace{-0.5cm}
	\end{figure}

	In Fig. \ref{visual_offline}, we provide the visual results of the optical flows generated by Spynet and our enhanced Spynet, the MV of VTM, and their corresponding warp frames. The MV of VTM is extracted in the luminance (Y) channel and interpolated on the chrominance (U, V) components to obtain a complete motion vector field, so only the visual results in the luminance channel are presented here. Under the guidance of the block-level MV searched by VTM, the enhanced optical flow can better recover the sharp motion boundaries and the regions with rich details visually.
	Compared with Spynet, the warp frames of enhanced Spynet have an average improvement of 1.15dB (33.23dB vs. 32.08dB) in JVET CTC test sequences~\cite{bossen16jvet}. The improvement in inter-frame prediction accuracy indicates that the offline enhancement can alleviate the domain gap between the synthetic data and the real-world videos to some degree.
	
	\subsection{End-to-End Training}
	After fine-tuning the pre-trained Spynet, we deploy it into DCVC, then train the whole video compression network in an end-to-end manner which is the same as DCVC. Thus, the training loss is as follows: 
	\begin{equation} 
		\label{rdloss}
		\setlength{\abovedisplayskip}{3pt}
		\setlength{\belowdisplayskip}{3pt}
		L = \lambda \cdot D + R = \lambda d(x_t, \hat{x}_t) + H(\hat{y}_t) + H(\hat{z}_t) + H(\hat{g}_t),
	\end{equation}
	where $\hat{y}_t=Q(y_t)$, $\hat{z}_t=Q(z_t)$, and $\hat{g}_t=Q(g_t)$. $Q(\cdot)$ represents the quantization operator. The term R in Eq. (\ref{rdloss}) denotes the number of bits used to encode the frame, and R is computed by adding up the number of bits $H(\hat{y}_t)$ and $H(\hat{z}_t)$ for encoding the latent features of motion information and $H(\hat{g}_t)$ for encoding the latent features of context. $d(x_t, \hat{x}_t)$ denotes the distortion between the input frame $x_t$ and the reconstruction frame $\hat{x}_t$. $\lambda$ is a hyperparameter that determines the trade-off between the number of bits $R$ and the distortion $D$. The MV of VTM is searched for the best trade-off between the bits cost and the MSE loss, so we only optimize our scheme with $D$ representing the MSE.
	
	\subsection{Online Enhancement} \label{on_enh}
	To further enhance the adaptivity of the optical flows and achieve a better compression performance, we propose the online enhancement on the optical flows. In the inference stage, we online optimize the latent features of the optical flows with a gradient descent-based algorithm minimizing the RD loss for the videos to be compressed. 
	
	\begin{figure}[t]
		\centering
		\includegraphics[width=\linewidth]{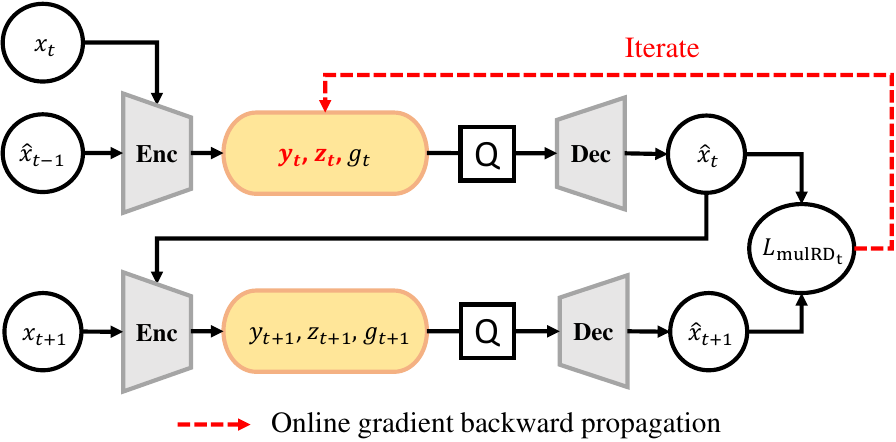}
		\caption{Overview of the three-frame online optimization.}
		\label{multi-frame}
		\vspace{-0.5cm}
	\end{figure}
	\begin{figure*}[t]
		\includegraphics[width=1\textwidth]{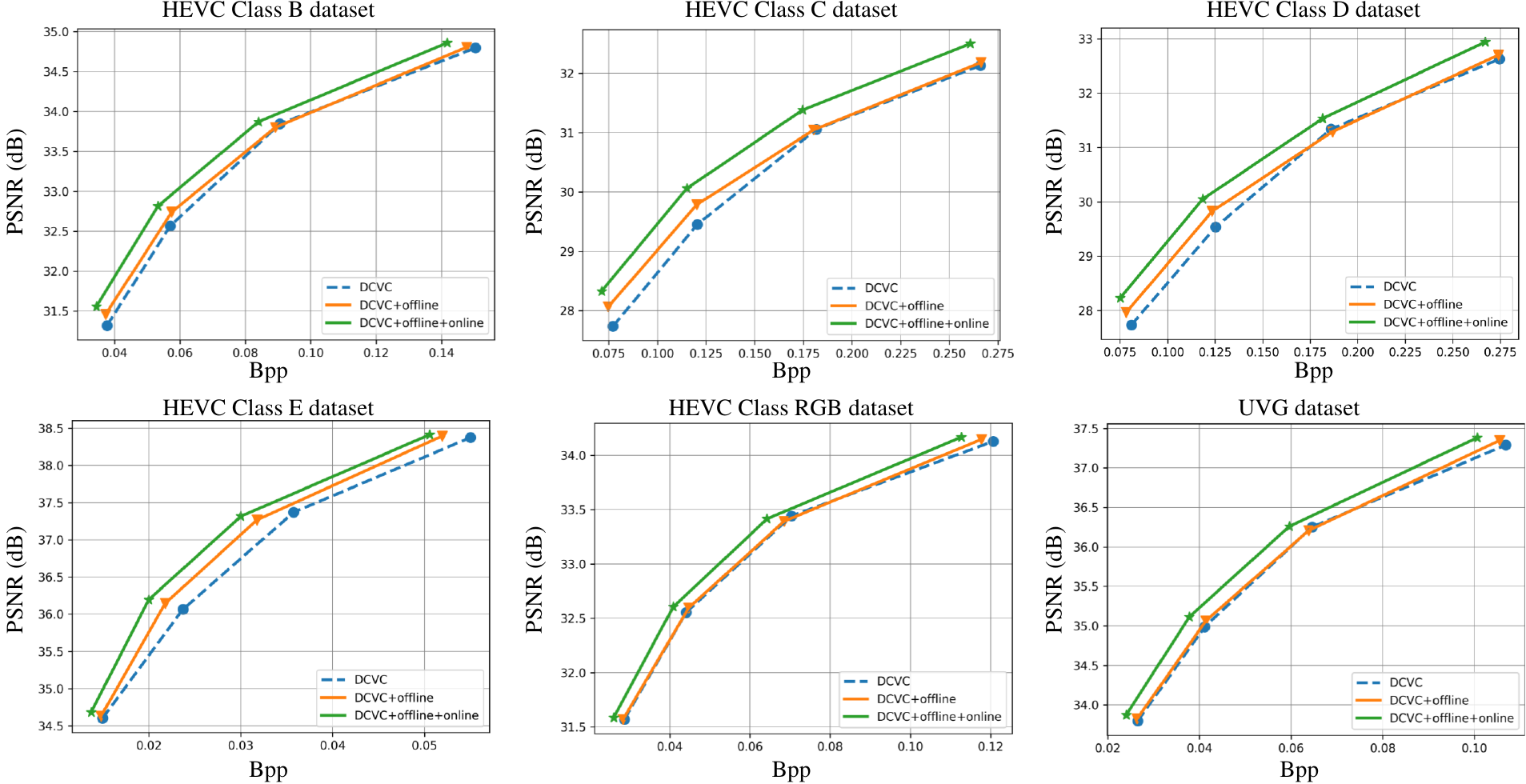}
		\caption{Rate-distortion curves of our scheme and DCVC on the HEVC Class B, Class C, Class D, Class E, Class RGB, and UVG datasets.}
		\label{main_curve}
	\end{figure*}
	
	{\bfseries Single-frame online optimization. }As shown in Fig. \ref{overview}(c), for the input frame $x_t$ and reference frame $\hat{x}_{t-1}$ in a group of pictures (GOP), we online update the latent features of the optical flows (MV feature $y_t$ and the MV hyperprior $z_t$) by a gradient descent-based algorithm 
	in the inference stage. After $N$ iterations, we obtain the latent features of the optical flows $\hat{y}^{op}_t$ and $\hat{z}^{op}_t$ which are optimal for the consecutive two frames $x_t$ and $\hat{x}_{t-1}$. Then, the latent features of context $\hat{g}_t$ and reconstruction frame $\hat{x}_t$ are generated by the $\hat{y}^{op}_t$ and $\hat{z}^{op}_t$, and we start to online update the latent features of the optical flows generated by the next input frame $x_{t+1}$ and the reference frame $\hat{x}_t$ in a GOP. 
	
	The pipeline of the optical flow latent updating algorithm is shown in Algorithm \ref{alg_sing}. Firstly, the initial latent features of the optical flows ${y_t}^{0}$ and ${z_t}^{0}$ are generated by the input frame $x_t$ and the reference frame $\hat{x}_{t-1}$. Secondly, the initial RD cost ${\hat{L}_{RD_t}}^{op}$ can be computed by feeding the initial latent features of the optical flows to the video decoder without gradient $\mathbf{Dec_I}$. Then, the latent features of the optical flows are online updated iteratively to minimize the RD loss of each iteration ${\tilde{L}_{RD_t}}^i$:
	\begin{equation} 
		\label{rdloss_online}
		\setlength{\abovedisplayskip}{3pt}
		\setlength{\belowdisplayskip}{3pt}
		{\tilde{L}_{RD_t}}^i = \lambda\cdot d(x_t, {\tilde{x}_{t}}^i) + H({\tilde{y}_t}^{i}) + H({\tilde{z}_t}^{i}) + H({\tilde{g}_{t}}^i),
	\end{equation}
	where ${{y}_t}^{i}$ denotes the MV feature of the current frame after $i$ steps of update, and ${{z}_t}^{i}$ denotes the MV hyperprior of the current frame after $i$ steps of update. Following the work~\cite{balle2016end}, we use adding uniform noise to approximate the rounding during training, ${\tilde{y}_t}^{i} = {{y}_t}^{i} + u$, ${\tilde{z}_t}^{i} = {{z}_t}^{i} + u$, and $u\sim\mathcal{U}(-0.5,0.5)$. The latent features of context ${\tilde{g}_{t}}^i$ and reconstruction frame ${\tilde{x}_{t}}^i$ are generated by feeding the latent features of the optical flows ${\tilde{y}_t}^{i}$ and ${\tilde{z}_t}^{i}$ to the video decoder with gradient $\mathbf{Dec_T}$ during the online optimization.
	
	The only difference between $\mathbf{Dec_T}$ and $\mathbf{Dec_I}$ lies in the quantization. To allow online optimization via gradient descent, the quantization in $\mathbf{Dec_T}$ is replaced by adding uniform noise, while the quantization in $\mathbf{Dec_I}$ is using rounding operation directly. During the updating iterations, the latent features of the optical flows are updated by minimizing the RD loss of each iteration ${\tilde{L}_{RD_t}}^i$, which is computed by sending latent features of the optical flows to $\mathbf{Dec_T}$.
	Then, the updated latent features of the optical flows are sent to $\mathbf{Dec_I}$ to compute the RD cost of each iteration ${\hat{L}_{RD_t}}^i$, and we only save the optimal latent features of the optical flows ${\hat{y}_t}^{op}$ and ${\hat{z}_t}^{op}$ which lead to the minimal RD cost ${\hat{L}_{RD_t}}^{op}$. 
	
	\begin{algorithm}[t]
		\caption{Optical Flow Latent Updating in the Inference Stage}
		\label{alg_sing}
		\begin{algorithmic}[1] 
			\STATE The MV encoder and hyperprior encoder $\mathbf{Enc_{MV}}$
			\STATE The video decoder with gradient $\mathbf{Dec_T}$
			\STATE The video decoder without gradient $\mathbf{Dec_I}$ 
			\STATE The input frame $x_t$ and the reference frame $\hat{x}_{t-1}$
			\STATE $\textit{N}$ represents the total updating times
			\STATE $\eta$ represents the step size (learning rate)
			\STATE $\lfloor \cdot \rceil$ represents the rouding operation and $u\sim\mathcal{U}(-0.5,0.5)$ 
			\STATE ${y_t}^{0}$, ${z_t}^{0}$ $\Leftarrow$ $\mathbf{Enc_{MV}}$($x_t$, $\hat{x}_{t-1}$)
			\STATE $ {\hat{y}_t}^{0},{\hat{z}_t}^{0}$ $\Leftarrow$ $\lfloor {y_t}^{0} \rceil, \lfloor{z_t}^{0}\rceil$
			\STATE ${\hat{x}_{t}}^0, {\hat{g}_{t}}^0$ $\Leftarrow$ $\mathbf{Dec_I}$(${\hat{y}_t}^{0}$, ${\hat{z}_t}^{0}$)
			\STATE $ {\hat{y}_t}^{op},{\hat{z}_t}^{op}$ $\Leftarrow$ $ {\hat{y}_t}^{0},{\hat{z}_t}^{0}$
			\STATE ${\hat{L}_{RD_t}}^{op}$ = $\lambda\cdot d(x_t, {\hat{x}_{t}}^0) + H( {\hat{y}_t}^{0}) + H( {\hat{z}_t}^{0}) + H({\hat{g}_{t}}^0)$
			\FOR {$ i = 0$; $ i < N $; $ i ++ $ }
			\STATE $ {\tilde{y}_t}^{i},{\tilde{z}_t}^{i}$ $\Leftarrow$ ${y_t}^{i}+u, {z_t}^{i}+u$
			\STATE ${\tilde{x}_{t}}^i, {\tilde{g}_{t}}^i$ $\Leftarrow$ $\mathbf{Dec_T}$(${\tilde{y}_t}^{i}, {\tilde{z}_t}^{i}$)
			\STATE ${\tilde{L}_{RD_t}}^i$ = $\lambda\cdot d(x_t, {\tilde{x}_{t}}^i) + H({\tilde{y}_t}^{i}) + H({\tilde{z}_t}^{i}) + H({\tilde{g}_{t}}^i)$
			\STATE ${y_t}^{i+1}$ $\Leftarrow$ ${y_t}^{i} - \eta\frac{\partial {\tilde{L}_{RD_t}}^i}{\partial {y_t}^{i}}$
			\STATE ${z_t}^{i+1}$ $\Leftarrow$ ${z_t}^{i} - \eta\frac{\partial {\tilde{L}_{RD_t}}^i}{\partial {z_t}^{i}}$
			\STATE ${\hat{y}_t}^{i+1},{\hat{z}_t}^{i+1}$ $\Leftarrow$ $\lfloor {y_t}^{i+1} \rceil, \lfloor{z_t}^{i+1}\rceil$
			\STATE ${\hat{x}_{t}}^{i+1}, {\hat{g}_{t}}^{i+1}$ $\Leftarrow$ $\mathbf{Dec_I}$(${\hat{y}_t}^{i+1}$, ${\hat{z}_t}^{i+1}$)
			\STATE ${\hat{L}_{RD_t}}^{i+1}$ = $\lambda\cdot d(x_t, {\hat{x}_{t}}^{i+1}) + H({\hat{y}_t}^{i+1}) + H({\hat{z}_t}^{i+1}) + H({\hat{g}_{t}}^{i+1})$
			\IF {${\hat{L}_{RD_t}}^{i+1} < {\hat{L}_{RD_t}}^{op}$}
			\STATE $ {\hat{y}_t}^{op},{\hat{z}_t}^{op}$ $\Leftarrow$ $ {\hat{y}_t}^{i+1},{\hat{z}_t}^{i+1}$
			\STATE ${\hat{L}_{RD_t}}^{op}$ $\Leftarrow$ ${\hat{L}_{RD_t}}^{i+1}$
			\ENDIF
			\ENDFOR
			\STATE ${\hat{x}_{t}}, {\hat{g}_{t}}$ $\Leftarrow$ $\mathbf{Dec_I}$ (${\hat{y}_t}^{op},{\hat{z}_t}^{op}$)		
		\end{algorithmic} 
	\end{algorithm}
	
	{\bfseries Multi-frame online optimization.} Considering the error propagation in deep video compression frameworks~\cite{lu2020content}, we further extend the single-frame online optimization algorithm to a multi-frame online optimization algorithm. We design a sliding-window-based online optimization algorithm to update the latent features of the optical flows by minimizing the multi-frame RD loss of each iteration ${\tilde{L}_{mulRD_t}}^i$ for all frames inside a window: 
	\begin{equation} 
		\label{rdloss_mul}
		\setlength{\abovedisplayskip}{3pt}
		\setlength{\belowdisplayskip}{3pt}
		{\tilde{L}_{mulRD_t}}^i = \sum_{j=t}^W \alpha_j[\lambda d(x_j, {\tilde{x}_{j}}^i) + H({\tilde{y}_j}^{i}) + H({\tilde{z}_j}^{i}) + H({\tilde{g}_{j}}^i)],
	\end{equation}
	where window size $W$ denotes the number of frames inside a window and $\alpha_j$ is a hyperparameter that determines the weight of RD loss for different frames.
	
	Specifically, the overview of three-frame online optimization is shown in Fig. \ref{multi-frame}. The consecutive three frames in a GOP $\hat{x}_{t-1}$, $x_t$, and $x_{t+1}$ are sent into the sliding window to update the latent features of the optical flows $y_t$ and $z_t$ iteratively minimizing multi-frame RD loss of each iteration ${\tilde{L}_{mulRD_t}}^i$. After $N$ iterations, we obtain the latent features of the optical flows $\hat{y}^{op}_t$ and $\hat{z}^{op}_t$ which are optimal for the consecutive three frames $\hat{x}_{t-1}$, $x_t$, and $x_{t+1}$, leading to the minimal multi-frame RD cost ${\hat{L}^{op}_{mulRD_t}}$. The reconstruction frame $\hat{x}_t$ is generated by the updated latent features $\hat{y}^{op}_t$ and $\hat{z}^{op}_t$, then the next consecutive three frames $\hat{x}_t$, $x_{t+1}$, and $x_{t+2}$ will be sent to the sliding window to update the next latent features of the optical flows. When the sliding window includes the last frame of the GOP, the window size $W$ will decrease by 1 until it equals to 2.
	The study in section~\ref{multi-frame_exp} shows the performance improvement on multi-frame online optimization with the increase of window size $W$.

	\section{Experimental Results}
	\subsection{Experimental Setup}
	{\bfseries Training Data.} We use BVI-DVC~\cite{ma2021bvi} dataset for fine-tuning Spynet~\cite{ranjan2017optical}. The BVI-DVC dataset contains 800 sequences at various spatial resolutions from 270p to 2160p. During the fine-tuning procedure, videos are cropped into non-overlapping 256 $\times$ 256 patches, and the motion vectors are extracted by VTM-10.0\footnote{\url{https://vcgit.hhi.fraunhofer.de/jvet/VVCSoftware_VTM/-/tree/VTM-10.0}} under the low-delay mode. The commonly-used Vimeo-90k~\cite{xue2019video} dataset is used for training DCVC in an end-to-end manner. During the training, all the videos are randomly cropped into 256 $\times$ 256 patches.
	
	{\bfseries Testing Data.} We use the JVET CTC test sequences~\cite{bossen16jvet} for evaluating the fine-tuning of Spynet. UVG~\cite{mercat2020uvg} and HEVC~\cite{bossen2013common} datasets are used for testing our scheme. The UVG dataset has 7 1080p sequences and the HEVC dataset contains 16 sequences including Class B, C, D, and E. In addition, we follow~\cite{li2022hybrid,sheng2022temporal,li2023neural}, and also evaluate on the HEVC RGB dataset~\cite{flynn2013common}. 
	
	{\bfseries Testing Conditions.} We test 96 frames for each video, and the intra period is set to 12 for each dataset. Besides, the same as DCVC, we use $Cheng2020Anchor$~\cite{cheng2020learned} implemented by CompressAI~\cite{begaint2020compressai} for intra-frame coding.
	
	\begin{figure*}[t]
		\includegraphics[width=1\textwidth]{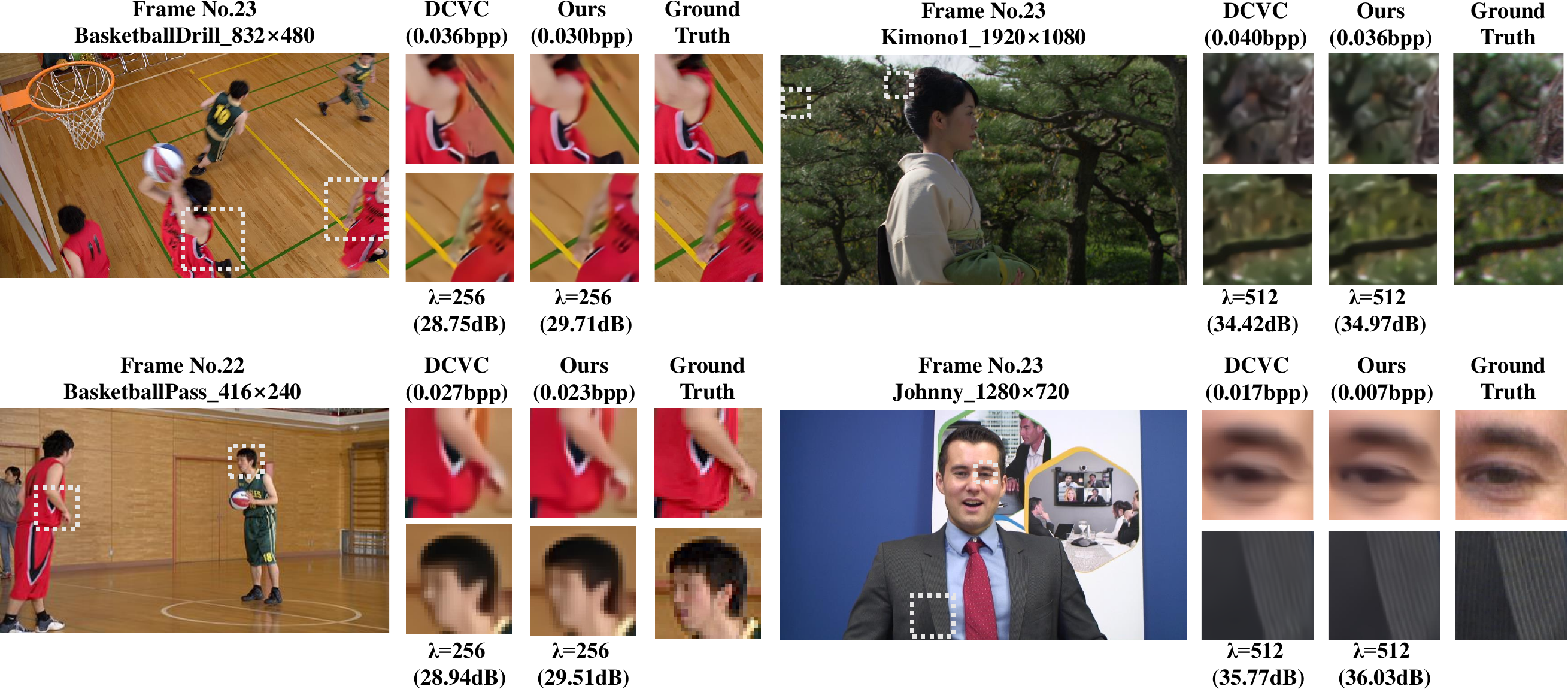}
		\caption{Reconstruction frame of DCVC and our scheme (DCVC + Offline + Online) and the ground truth. The four reconstruction frames are in different sequences from HEVC dataset.}
		\label{visual_recon}
	\end{figure*}
	
	{\bfseries Implementation Details.} Our scheme includes three training stages, which consist of the fine-tuning of Spynet, offline training of DCVC, and online optimization of DCVC with the enhanced Spynet. In the first stage, we set $\lambda_{ME}$ to 100, and fine-tune the Spynet using the extracted MV for 1,000,000 iterations. In the second stage, we deploy the enhanced Spynet into DCVC and train the whole video compression network for 5,000,000 iterations with different $\lambda$ values (256, 512, 1024, 2048) until converge. The anchor DCVC is also trained for the same steps. Finally, we set the updating times $N$ in Algorithm \ref{alg_sing} to 1500 according to the ablation study in~\ref{abl}, then online update the latent features of the optical flows for 1500 iterations. The initial learning rate for the first two steps is 1e-4, then decrease to 5e-5 at the 800,000th iteration and 4,000,000th iteration respectively. The initial learning rate for online optimization is 5e-3, which is decreased by 50\% at the 1200th iteration. The Adam optimizer~\cite{kingma2014adam} is used, and the batch size is set to 16 for the first training stage and 4 for the second training stage. We train Spynet on a single NVIDIA 2080TI GPU for about 5 days and DCVC on four NVIDIA 2080TI GPUs for about 7 days. Besides, the online optimization experiments are conducted on NVIDIA 3090 GPU. 
	
	\subsection{Comparisons with Baseline Method}
	Fig.~\ref{main_curve} shows RD curves on HEVC Class B, Class C, Class D, Class E, Class RGB, and UVG datasets. Our baseline scheme is DCVC, and it's obvious that DCVC with the offline and online enhancement on the optical flows can outperform DCVC in all rate points. Besides, both the offline and online enhancement on the optical flows don't change the network structure of DCVC and only optimize the encoder side of DCVC, leading to no increase in the model size or computational complexity on the decoder side. The proposed offline and online enhancement together achieves an average of 12.8\% bitrate saving on all testing datasets over DCVC. The offline enhancement can achieve an average of 3.9\% bitrate saving on all testing datasets over DCVC, which verifies that the offline enhancement can provide a more appropriate initial point for the end-to-end optimization in the deep video compression network. 
	
	
	In Fig~\ref{visual_recon}, we present visual results of the reconstruction frames of DCVC and our scheme across different sequences. With the offline and online enhancement on the optical flows, our scheme can achieve higher quality reconstruction, retaining more details in the boundaries of the motion and the regions with rich texture, while using fewer bits than DCVC. 
	
	\begin{table}
		\centering
		\caption{Effectiveness of the offline and online enhancement. BD-Rate(\%) comparison for PSNR. The anchor is DCVC. Negative values in BDBR represent the bitrate saving.}
		\label{ablation_on_off}
		\scalebox{0.85}{\begin{tabular}{ccccccccc}
				\Xhline{1px}
				Offline & Online & B     & C     & D     & E     & RGB  & UVG   & Average \\ \Xhline{1px}
				×       & ×      & 0.0   & 0.0   & 0.0   & 0.0   & 0.0  & 0.0   & 0.0     \\ \hline
				\checkmark       & ×      & -3.0  & -5.9  & -4.4  & -7.9  & -0.7 & -1.3  & -3.9    \\ \hline
				×       & \checkmark      & -10.7 & -14.3 & -11.1 & -9.0  & -8.5 & -10.1 & -10.6   \\ \hline
				\checkmark       & \checkmark      & -12.0 & -17.1 & -13.1 & -15.3 & -8.8 & -10.5 & -12.8   \\ \Xhline{1px}
		\end{tabular}}
	\end{table}

	\begin{table}[h]
		\caption{BD-Rate(\%) comparison for PSNR, Encoding time $ENC_T$, and Decoding time $DEC_T$ for different updating times $U$ in the online enhancement. The anchor is DCVC + Offline ($N = 0$).}
		\label{tab:update_time}
		\scalebox{0.85}{
			\begin{tabular}{ccccccl}	
				\Xhline{1px}
				$U$ &C & D &$ENC_T$ C(s) & $DEC_T$ C(s) &$ ENC_T$ D(s) & $DEC_T$ D(s)\\ \Xhline{1px}
				0 & 0.0& 0.0 &2.71 & 6.94 &0.70 & 1.91\\ \hline
				100 & -6.1 & -5.1 & 28.15 & 6.84 & 10.42 & 1.90 \\ \hline
				500 & -9.6 & -7.9 & 132.78 & 6.95 & 48.99 & 1.87\\ \hline
				1000 & -10.8 & -8.6 & 269.20 & 6.73 & 92.58 & 1.89\\ \hline
				1500 & -11.2 & -8.7 & 388.73 & 6.86 & 141.03 & 1.91\\ \hline
				2000 & -11.5 & -9.1 & 530.10 & 6.84 & 190.64 & 1.89\\ \hline
				2500 & -11.6 & -9.2 & 674.54 & 6.89 & 239.05 & 1.88\\ \Xhline{1px}
			\end{tabular}
		}
	\end{table}
	
	\subsection{Ablation Study} \label{abl}
	{\bfseries Effectiveness of Offline and Online Enhancement.} To verify the effectiveness of the offline and online enhancement on the optical flows, we compare the compression performance of the baseline scheme (DCVC) with or without the enhancement. We report the BD-rate~\cite{bjontegaard2001calculation} results in Table~\ref{ablation_on_off}. For online enhancement, the updating times are set to 1500.  
	From the comparison results, we find that the offline enhancement on the optical flows brings 3.9\% bitrate saving and the online enhancement brings 10.6\% bitrate saving.  With both offline and online enhancement, 12.8\% bitrate saving is achieved. The experimental results indicate that our offline enhancement on the optical flows can also provide a more appropriate initial point for the online optimization.
	
\begin{figure}[t]
		\centering
		\includegraphics[width=\linewidth]{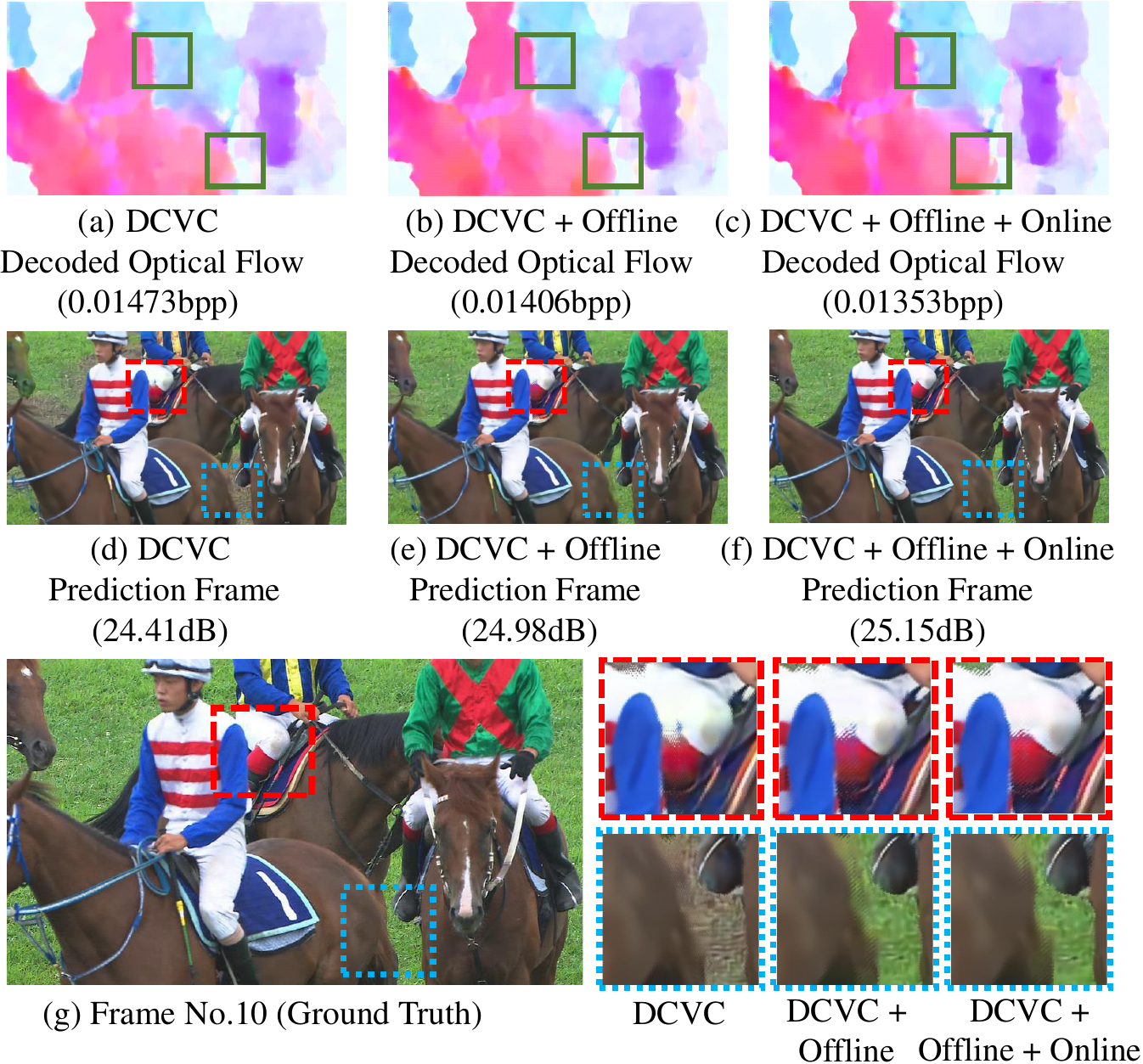}
		\caption{Visual comparison of the decoded optical flows and the corresponding prediction frames obtained by DCVC, DCVC + Offline, and DCVC + Offline + Online with $\lambda = 256$. We also provide (g) the ground truth which is the 10th frame in RaceHorses\_832$\times$480 as a reference.}
		\label{visual_abl}
	\end{figure}

	In Fig.~\ref{visual_abl} (a) - (f), we provide the visual results of the decoded optical flows and their corresponding prediction frames. It is observed that with our offline enhancement on the optical flows, accurate motion-compensated results can be generated in the homogeneous region, resulting in a 0.57dB improvement with fewer bits cost over DCVC. The online enhancement on the optical flows further improves the decoded optical flow in areas close to motion boundaries, which has further achieved 0.17dB improvement with even fewer bits cost. Compared with the raw frame shown in Fig.~\ref{visual_abl} (g), the prediction frames can recover increasingly more details in the raw frame by adopting the offline and online enhancement on the optical flows.
	
	{\bfseries Influence on updating times of online enhancement.} To study the influence of the total updating times in online enhancement, we change the updating times from 100 to 2500. For simplification, we only use HEVC Class C and Class D datasets to explore the reasonable updating times considering the trade-off between compression performance and encoding time. The anchor is DCVC with the offline enhancement on the optical flows (DCVC + Offline). Table \ref{tab:update_time} reports the BD-rate~\cite{bjontegaard2001calculation}, which indicates that the RD performance is improved as the updating times $N$ increases. To balance the trade-off between the compression performance and encoding time complexity, in this paper, we set the updating times $N$ to 1500 for online enhancement. Besides, the decoding time in Table \ref{tab:update_time} demonstrates that our proposed method doesn't increase the computational complexity on the decoder side.
	
	\subsection{Multi-frame Online Optimization} \label{multi-frame_exp}
	In this paper, we adopt the single-frame online optimization in the inference stage to improve the compression performance and achieve content-adaptive encoding. Besides, we also provide the compression results adopting the multi-frame online optimization which updates the latent features of the optical flows by minimizing multi-frame RD loss. 
	We wish to explore the potential of multi-frame online optimization on the motion information with a limited number of frames in a GOP.
	
	For simplification, we only conduct experiments on HEVC Class C and Class D datasets. we set the DCVC with offline and online enhancement on the optical flows (DCVC + Offline + Online) as the anchor, which adopts the single-frame online optimization. The single-frame online optimization is the same as setting the window size $W$ in Eq. (\ref{rdloss_mul}) to 2. The hyperparameters $\alpha_0$, $\alpha_1$, $\alpha_2$, and $\alpha_3$ in Eq. (\ref{rdloss_mul}) are set to 1, 0.5, 0.2, and 0.1 respectively. Table \ref{tab:update_frames_gop12} reports the BD-rate~\cite{bjontegaard2001calculation} with updating times $N$ set to 2000. We compare the compression performance, encoding time, and decoding time of DCVC with multi-frame online enhancement on the optical flows with window size $W$ set from 2 to 5 in HEVC Class C and Class D datasets. Table~\ref{tab:update_frames_gop12} shows that increasing the window size cannot improve the compression ratio greatly, but the encoding time has increased a lot when the window size exceeds 2. In this paper, we currently adopt the single-frame online optimization. 
	
	\begin{table}[h]
		\caption{BD-Rate(\%) comparison for PSNR, Encoding time $ENC_T$, and Decoding time $DEC_T$ for different window size $W$ in the multi-frame online enhancement. The anchor is DCVC + Offline + Online, which adopts the single-frame online optimization ($W = 2$).}
		\label{tab:update_frames_gop12}
		\scalebox{0.85}{
			\begin{tabular}{ccccccl}	
				\Xhline{1px}
				$W$& C& D &$ENC_T$ C(s) & $DEC_T$ C(s) &$ENC_T$ D(s) & $DEC_T$ D(s)\\ \Xhline{1px}
				2 & 0.0 & 0.0 & 518.25 & 6.84 & 187.82 & 1.89\\ \hline
				3 & -0.5 & -0.4 & 1631.35 & 6.84 & 546.99 & 1.88\\ \hline
				4 & -0.7 & -0.7 & 2187.58 & 6.82 & 683.04 & 1.88\\ \hline
				5 & -0.8 & -0.8 & 2706.39 & 6.87 & 874.56 & 1.86\\ \Xhline{1px}
			\end{tabular}
		}
	\end{table}

	\section{Conclusion}
	In this paper, we have proposed an offline and online enhancement on the optical flows to better estimate and utilize the motion information in the deep video compression network. Specifically, in the offline enhancement, we fine-tune the optical flow estimation network with the MV of VTM, which is searched for the best RD performance on real-world videos. In the online enhancement, we online update the latent features of the optical flows under the RD metric for different coding sequences in the inference stage. Our scheme can effectively improve the compression performance without increasing the model size or computational complexity on the decoder side. The experimental results show that our scheme can outperform our baseline scheme DCVC in terms of PSNR by 12.8\% under the same configuration. 
	
	It is worth noting that our scheme is a plug-and-play mechanism that can be easily integrated into any deep video compression framework with the same motion estimation and MV encoder. We believe that our scheme could help researchers to further explore what is a better motion information and how to compress it effectively in deep video compression frameworks. 
	
	\newpage
	
	\bibliographystyle{IEEEtran}
	\bibliography{refs.bib}

\end{document}